# Overview of the MedHopQA track at BioCreative IX: track description, participation and evaluation of systems for multi-hop medical question answering

Rezarta Islamaj[1], Joey Chan[1,2], Robert Leaman[1], Jongmyung Jung[3], Hyeongsoon Hwang[3], Quoc-An Nguyen[4], Hoang-Quynh Le[4], Harikrishnan Gurushankar Saisudha[5], Ganesh Chandrasekar[5], Rustam R. Taktashov[6], Nadezhda Yu. Bizyukova[6], Sofia I. R. Conceição[7], Paulo R. C. Lopes[7], Reem Abdel Salam[8,10], Mary Adewunmi[9,10] and Zhiyong Lu[1]

[1]National Library of Medicine (NLM), National Institutes of Health (NIH), MD, 20894, Bethesda, USA
[2]University of Illinois at Urbana Champaign
[3]Korea University
[4]VNU University of Engineering and Technology, Hanoi, Vietnam
[5]Concordia University, Montreal, QC, CA
[6] Institute of Biomedical Chemistry (IBMC), 10 bld. 8, Pogodinskaya str., 119121 Moscow, Russia
[7]LASIGE, Departamento de Informática, Faculdade de Ciências, Universidade de Lisboa, 1749-016 Lisbon, Portugal
[8]Faculty of Engineering, Computer Engineering Department Cairo University
[9]Menzies School of Health Research, Charles Darwin University, NT, Australia
[10]CaresAI, Australia

*Corresponding author: E-mail: Zhiyong.Lu@nih.gov

## Abstract

Multi-hop question answering (QA) remains a significant challenge in the biomedical domain, requiring systems to integrate information across multiple sources to answer complex questions. To address this problem, the BioCreative IX MedHopQA shared task was designed to benchmark in multi-hop reasoning for large language models (LLMs). We developed a novel dataset of 1,000 challenging QA pairs spanning diseases, genes, and chemicals, with particular emphasis on rare diseases. Each question was constructed to require two-hop reasoning through the integration of information from two distinct Wikipedia pages.

The challenge attracted 48 submissions from 13 teams. Systems were evaluated using both surface string comparison and conceptual accuracy (MedCPT score). The results showed a substantial performance gap between baseline LLMs and enhanced systems. The top-ranked submission achieved an 89.30% F1 score on the MedCPT metric and an 87.30% exact match (EM) score, compared with 67.40% and 60.20%, respectively, for the zero-shot baseline. A central finding of the challenge was that retrieval- augmented generation (RAG) and related retrieval-based strategies were critical for strong performance. In addition, concept-level evaluation improved answer assessment when correct responses differed in surface form. The MedHopQA dataset is publicly available to support continued progress in this important area.

Challenge materials: https://www.ncbi.nlm.nih.gov/research/bionlp/medhopqa and benchmark https://www.codabench.org/competitions/7609/

## Introduction

For two decades, BioCreative challenges(1-3) have played a central role in advancing the foundational methods of biomedical text mining. In that tradition, the introduction of the MedHopQA (Medical Multi-Hop Question Answering) task in BioCreative IX marks a timely shift from information extraction toward information synthesis and reasoning, driven by the recent maturation of large language models (LLMs).

Earlier BioCreative challenges focused on the foundational problem of creating algorithms to structuring the biomedical literature(4-6). Article triage, ranking, and classification tasks aimed to identify documents containing biomedically relevant information(7). Named entity recognition (NER) tasks focused on identifying core biological entities(8-10), and relation extraction tasks focused on mapping relationships among them(7,11-13). Together, these efforts supported biocuration tasks essential for populating resources such as Gene Ontology, BioGRID, and CTD. The central objective was to extract discrete, verifiable facts from text and convert them into structured, machine-readable format.

The landscape of artificial intelligence (AI) has been fundamentally altered by the arrival of powerful LLMs and expanded the scope of what biomedical text-mining systems can attempt (14-16). These models are designed not only to identify information, but also to synthesize and reason across it. This capability is particularly relevant in biomedicine, where the literature has expanded beyond what any researcher or clinician can realistically track and where many real-world questions require integrating evidence distributed across multiple sources.  For example, answering a question about how a gene variant may influence response to a therapy may require linking the variant to its gene, the gene to its protein product, and the protein to the therapy’s mechanism of action. Biomedical information needs are therefore often compositional, requiring systems to chain inference across dispersed evidence.

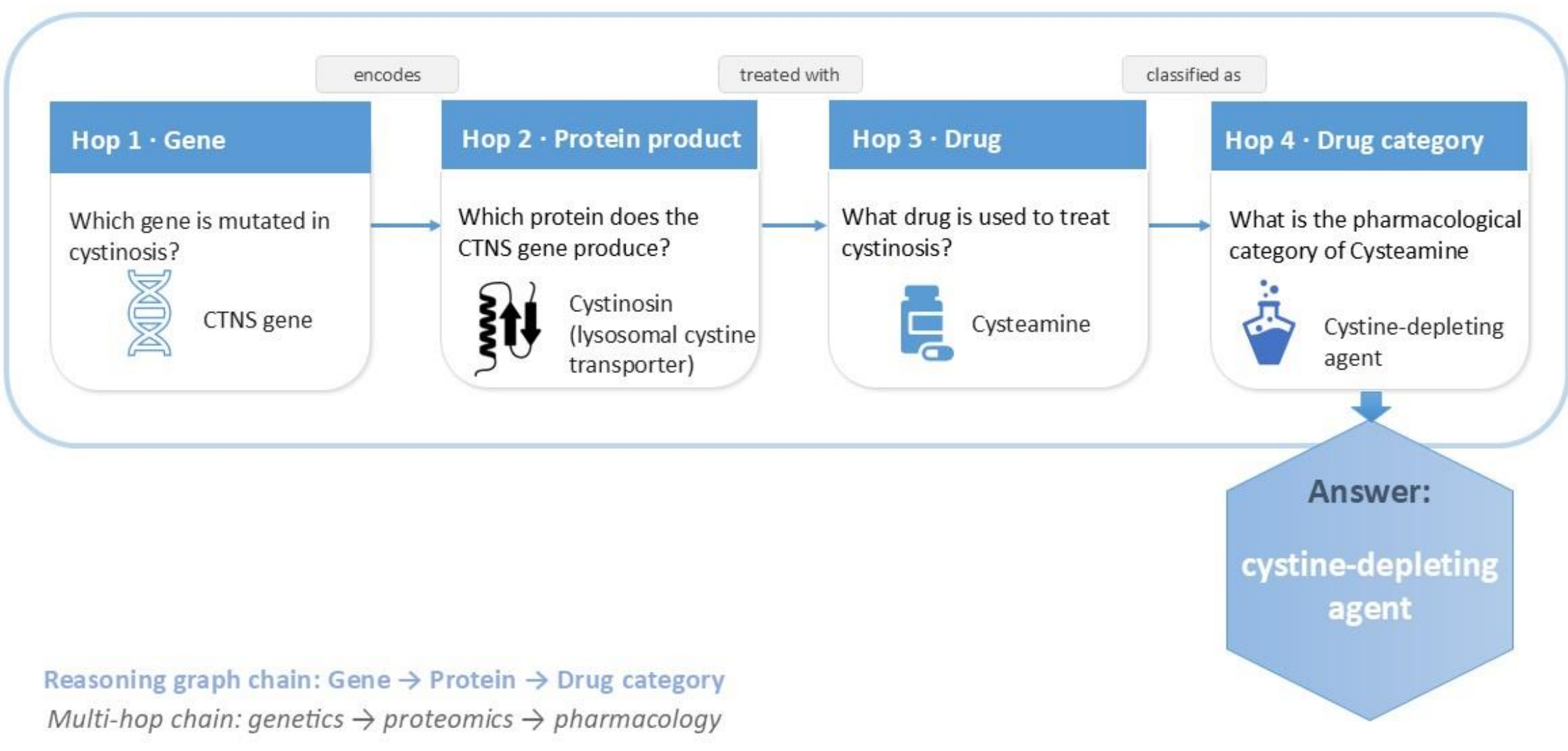

*Figure 1 A depiction of a multi-hop question and the reasoning process required to answer correctly.*

As shown in Figure 1, the MedHopQA challenge is designed to evaluate the multi-hop reasoning ability of biomedical question-answering systems (17). Whereas earlier BioCreative challenges emphasized systems that could identify, extract, and organize information, MedHopQA evaluates whether systems can integrate evidence and reason across sources.

Biomedical question answering (QA) aims to support efficient evidence retrieval, hypothesis generation, and decision support(18). Shared-task benchmarks have been central to progress in this area by standardizing datasets, evaluation protocols, and reporting practices. For example, BioASQ (19) combines biomedical QA with document and snippet retrieval and expert assessment, thereby encouraging evidence-grounded answering in realistic settings. Other datasets cover complementary QA settings, including research-article QA, such as PubMedQA (20), and professional exam-style multiple-choice QA, such as MedQA (21)) and MedMCQA (22). Together, these resources span a broad range of biomedical reasoning styles and difficulty levels. Table 1 summarizes biomedical and medical QA datasets and selected key characteristics.

Recent advances in large language models (LLMs) have led to substantial gains on many of these benchmarks. However, aggregate evaluations such as MultiMedQA have shown that strong benchmark performance does not necessarily translate into reliable clinical behavior, underscoring the need to evaluate dimensions beyond accuracy, including factuality and potential harm (23).

In addition, as shown in Table 1, many widely reported QA results are driven by single-hop questions that can be answered from a single passage or localized evidence in the document. In contrast, real biomedical information needs are often compositional and require multi-hop inference across dispersed evidence. For example, answering a clinically meaningful question may require linking gene-disease associations, mechanistic or pathway-level knowledge, and chemical or therapeutic properties to justify a possible intervention. Benchmarks that explicitly require multi-document evidence integration provide a useful model, but biomedical multi-hop resources remain relatively limited and have not yet become a standard basis for evaluation (24).

*Table 1 Summary of community challenges on biomedical QA benchmarks and related tasks.*

| **Resource / Task** | **Primary focus** | **Typical source text** | **Answer format** | **Evidence grounding expected?** | **Multi-hop reasoning?** |
|---|---|---|---|---|---|
| **BioASQ (Task B: biomedical semantic QA)(19)** | Biomedical QA + retrieval (IR + exact + ideal answers) | PubMed abstracts/articles (with retrieval component) | Exact answer + “ideal” summary | Yes (documents/ snippets) | Sometimes, often single-hop / localized evidence |
| **PubMedQA (20)** | Research-abstract QA (reasoning over a single abstract) | PubMed abstracts | Yes/No/Maybe | Implicit (in the abstract) | single-hop / single-document |
| **MedQA (USMLE-style) (21)** | Medical exam-style reasoning and knowledge | Question stem + options (external knowledge) | Multiple choice | Not required | Generally multi-step; not explicitly “multi-hop chaining” |
| **MedMCQA (22)** | Large-scale medical entrance exam MCQs | Question stem + options | Multiple choice | Not required | Often multi-step, but not explicitly multi-document |

| **MultiMedQA (suite)**[1] | Aggregate evaluation of medical QA + human-judged safety/factuality aspects | Mix of tasks/datasets (suite) | Varies (MCQ / free-form / etc.) | human evaluation | Depends; not solely multi-hop |
|---|---|---|---|---|---|
| **QAngaroo / MedHop** [2] | Multi-hop reading comprehension across documents | Multiple documents/ passages | Short answer | combine evidence across docs | Yes — explicit multi-hop |
| **emrQA(25)** | Clinical QA grounded in EHR notes | Clinical notes (EHR-style) | Short spans / structured answers | Grounded in the note | Often single-document; can be compositional |
| **MEDIQA (shared task)(26)** | Medical QA ecosystem + entailment / inference (RQE, NLI, QA) | Clinical/consumer health text (varies by subtask) | Varies by subtask | Often expected | Not primarily; |
| **TREC Clinical Decision Support (CDS) (27)** | Retrieval for clinical decision support (QA-adjacent) | Patient record + literature corpus | Ranked documents/snippets | Yes (retrieval-centric evidence) | Can require multi-step info seeking; evaluated as retrieval |

Compared with previous challenges listed in Table 1, MedHopQA makes several important contributions. Many earlier shared tasks were built on benchmark datasets in which answers can often be derived from a single abstract, snippet, or other localized evidence. However, biomedical reasoning often needs multiple inference steps across different sources, which MedHopQA is designed to explicitly require. Each question is curated from two interconnected Wikipedia pages containing disease-centered information, such as signs and symptoms, genetics, and treatments. This design also aligns well with current LLM-based QA paradigms, including retrieval-augmented and agentic workflows.

MedHopQA also improves evaluation fidelity by supporting concept-level equivalence, allowing semantically identical answer variants to be treated as correct. As a BioCreative IX shared task, it further provides a community benchmark for developing and evaluating robust biomedical multi-hop QA systems.

In this manuscript, we provide an overview of the MedHopQA shared task at BioCreative IX. We first describe the dataset and its baseline evaluation framework, then summarize selected participating systems and compare their design strategies. We present the evaluation results and discuss the system characteristics associated with stronger performance. The MedHopQA dataset and other challenge materials are available at (https://www.ncbi.nlm.nih.gov/research/bionlp/medhopqa) and (https://www.codabench.org/competitions/7609/). The MedHopQA evaluation remains open to the public through Codabench, and the benchmark has continued to attractCodabench interest beyond the formal challenge period.

[1] MultiMedQA/README.md at main · monk1337/MultiMedQA · GitHub
[2] QAngaroo -- Reading Comprehension with Multiple Hops

## Materials and methods

### Dataset description

The MedHopQA dataset focuses on questions involving diseases, genes, and chemicals, with particular emphasis on rare diseases. This focus ensures that the questions are both medically relevant and challenging, thereby testing the limits of current AI systems. The dataset contains 1,000 question-answer pairs. For official evaluation, these questions are embedded within a larger set of 10,000 questions to support robust and unbiased assessment.

MedHopQA was designed to evaluate question answering tasks that require multi-step reasoning across biomedical entities. Unlike many QA benchmarks that can be solved through single-hop retrieval or isolated fact lookup, MedHopQA questions require systems to decompose the question into intermediate steps, maintain context across those steps, and integrate its findings into a coherent final answer. Because key biomedical relationships are often distributed across multiple sources, successful performance requires compositional reasoning over dispersed evidence.

The questions and answers are derived from publicly available information, specifically, Wikipedia. This choice was intentional: it allows evaluation of whether models can reason over information they have likely encountered during training, rather than simply assessing recall of obscure sources. Each question is constructed from two interconnected Wikipedia pages, targeting and targets disease-relevant knowledge, such as signs and symptoms, genetics, and treatments. The dataset includes both entity-based answers, such as disease names or genes, and yes/no questions. Figure 2 summarizes the distribution of answer types. The answer category “Other” includes less common types such as person name, date, cell type, and procedure or treatment type.

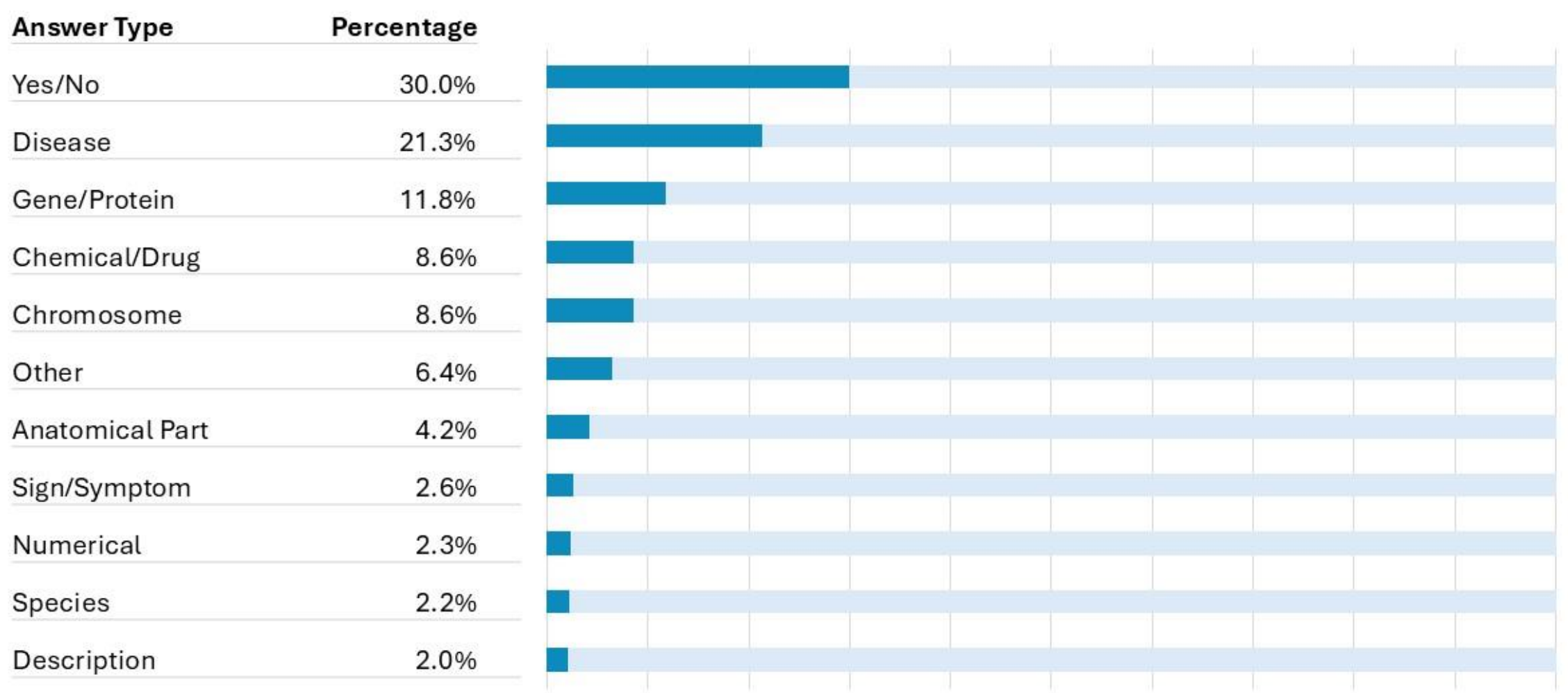

*Figure 2 The distribution of answer type categories in the MedHopQA corpus.*

### Track description and participation

The MedHopQA shared task was announced in March 2025 as part of the BioCreative IX workshop held in conjunction with the International Joint Conference on Artificial Intelligence (IJCAI) 2025 (Figure 3). The task was hosted on the Codabench platform and attracted registrations from 42 international teams. Organizers provided the evaluation methodology together with sample questions and answers through both the official MedHopQA and Codabench websites.

In accordance with the IJCAI conference timeline, the formal evaluation period for BioCreative IX took place from May 27 to June 1, 2025. During this period, participants were allowed to submit up to five system runs, resulting in 48 submissions. In response to participant requests for additional time, the organizers also offered an extended unofficial testing period from June 10 to June 18, during which 19 additional submissions were received. After the challenge, participating teams prepared system papers describing their methods for peer review and publication in the BioCreative IX proceedings (28). The official results and findings were presented at the BioCreative IX workshop on August 17-18, 2025.

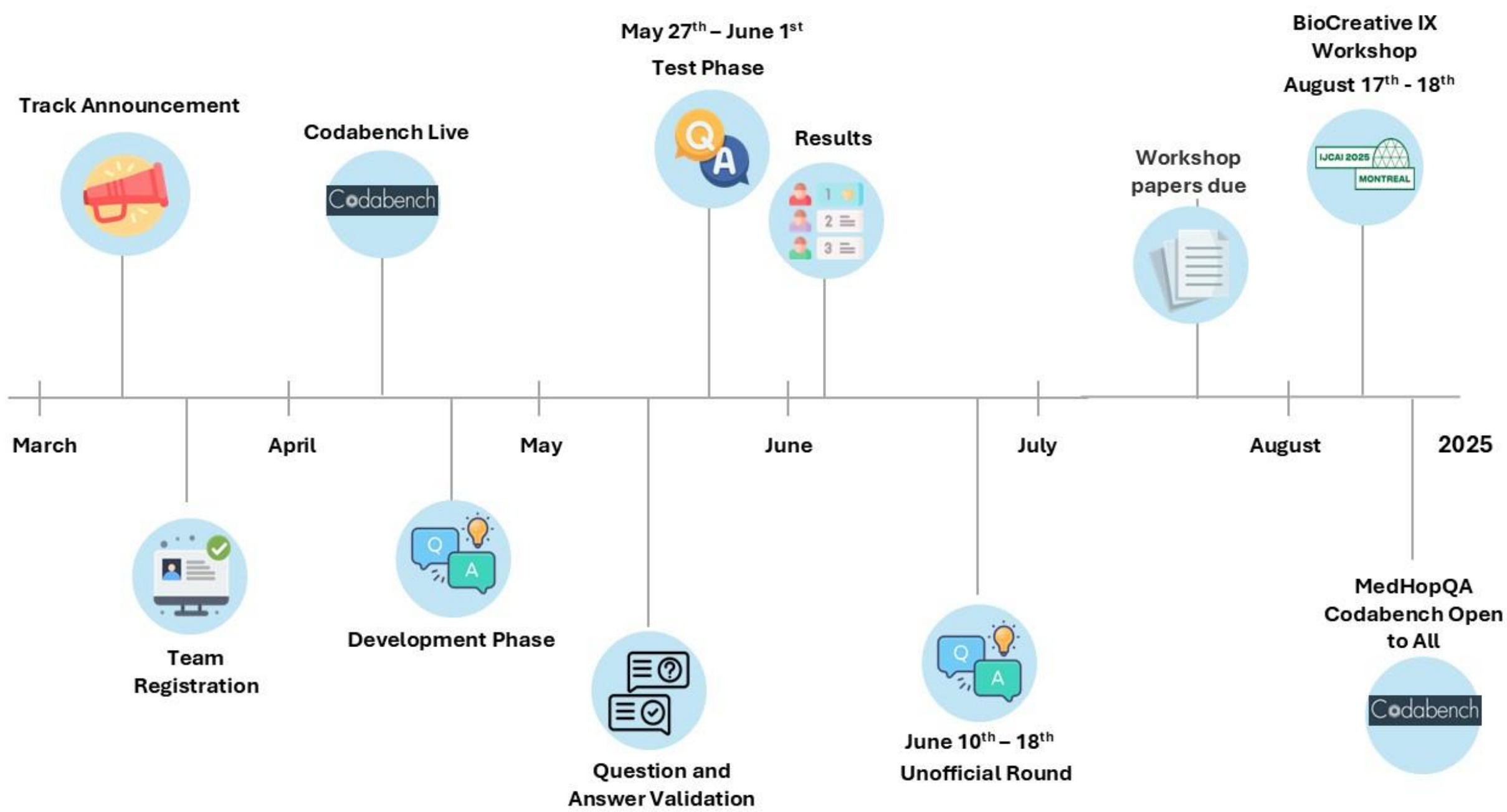

*Figure 3 The MedHopQA shared task timeline, from announcement to workshop, illustrating task preparation and communication with participants from March to August 2025.*

## MedHopQA evaluation methods

### The challenge of evaluating LLM-based models

Evaluating LLMs practical when a community shared task presents substantial practical challenges, particularly on public platforms such as Codabench. A successful challenge requires a robust and scalable evaluation backend that can process many submissions, execute evaluation code efficiently, and return timely feedback to participants. For LLM-based systems, these requirements are complicated by the size and computational demands of modern models, as well as the variability and nuance of their outputs. Model files may be several gigabytes in size, making transfer slow and error-prone, while loading them the evaluation server would further increase runtime and system complexity. Many LLMs may contain billions of parameters and require substantial memory and specialized hardware for inference, whereas public platforms such as Kaggle and Codabench impose strict limits on computational resources available for submission. In addition, custom LLM pipelines often depend on specific library and framework versions, making it difficult to guarantee compatibility between participant environments and the platform's infrastructure. For these reasons, requiring participants to submit full models for evaluation is impractical. However, preserving evaluation integrity requires the test set to be handled carefully to minimize the risk of leakage.

To address these constraints, the organizers embedded the 1,000 evaluation questions within a larger collection of decoy questions, yielding a total of 10,000 questions. Participants ran their systems locally and submitted only their predictions, which could then be uploaded to Codabench for lightweight, rapid evaluation.

**The challenge of semantic precision in medical LLM evaluation**

The medical domain introduces additional evaluation challenges (18). Conventional QA metrics often rely on lexical overlap, using measures such as ROUGE, BLEU, or F-score computed over surface forms. Although such metrics are useful in general-domain settings, they are insufficient for biomedicine, where subtle textual differences may correspond to clinically distinct concepts. For example, "*Diabetes Mellitus Type 1*" and "*Diabetes Mellitus Type 2*" are highly similar strings but represent distinct conditions with important implications for patient care. A lexical-overlap metric may therefore overestimate correctness when the underlying concept is wrong.

**A two-tiered evaluation strategy**

To address this limitation, we augmented the MedHopQA dataset with a comprehensive thesaurus of clinical synonyms and accepted answer variants for each question. These answer sets were manually compiled using biomedical knowledge resources such as MeSH, NCBI Gene, NCBI Taxonomy, Disease Ontology, Mondo, and Cell Ontology. Because biomedical ontologies differ in scope, structure, and classification principles, terms treated as equivalent in one resource are not necessarily interchangeable in all settings (29). Individual review was necessary to ensure that synonym equivalence was appropriate in context. For MedHopQA, the relevant context for equivalence was the question itself.

On this basis, we developed a two-tiered evaluation strategy for the MedHopQA task:

1. **String matching with synonym lexicon**: For the automated Codabench leaderboard, we implemented a lightweight string-matching method that we term "Lexical Match." This was not a simple exact-match comparison. Instead, the script scored a submitted answer according to its similarity to any semantically equivalent answer in the curated lexicon while remaining computationally efficient.

2. Concept-level validation: As a secondary and more semantically robust evaluation, we used MedCPT (30), a specialized biomedical model, to perform concept-level matching. Because of its computational requirements, this analysis was conducted locally and the results were communicated to participants separately.

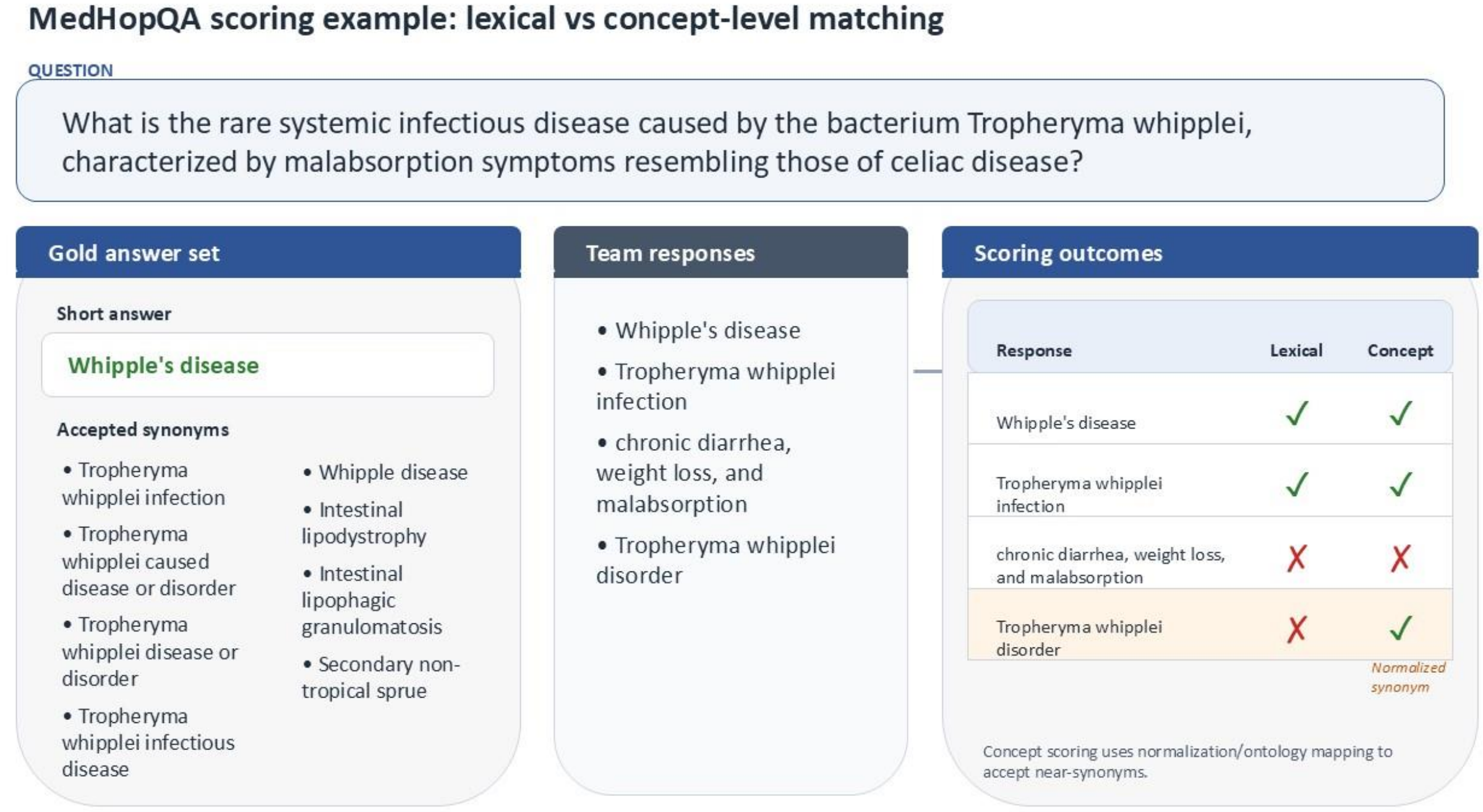

*Figure 4 MedHopQA lexical matching (string matching with synonym lexicon) and concept-level evaluation*

The curated synonym lexicon produced a strong correlation between the efficient lexical match metric and the deeper concept-level analysis. This indicates that the lexical match approach serves as a reliable proxy for semantic correctness within the automated evaluation pipeline. As expected, concept-level scores were generally higher because they capture a broader range of semantic equivalences. Figure 4 illustrates an example from the dataset, showing responses from different teams together with the scores assigned by the lexical match and concept-level evaluations.

### The MedHopQA baseline system

The baseline system used a zero-shot GPT-4o model with temperature set to zero and a 15-token output limit. The system used the prompt: "*You are a helpful assistant. Please give a short concise answer for the question consisting of one word or phrase.*" Future work will extend this evaluation framework to support longer, explanatory answers using an agent-based evaluation pipeline.

## Submitted systems

*Table 2 Summary of the systems that submitted a workshop paper. For each system we list the team name, country, the LLMs used, a brief description of their retrieval strategy, the knowledge sources incorporated, and the number of runs submitted for evaluation, including runs submitted after the official test phase.*

| Team | LLM(s) Used | Retrieval/Generation Strategy | Knowledge Sources | Official (add'l) runs |
|---|---|---|---|---|
| DMIS* Lab(31) (South Korea) | GPT-4o, GPT-3, Gemini 2.5 Pro | Three-arm pipeline: (1) Query reformulation → local Wikipedia retrieval (BM25 + MedCPT) → few-shot answering; (2) CoT rationale to create subqueries → same retrieval; (3) Web-search augmented: Google Search evidence → Gemini 2.5 Pro step-by-step reasoning | Wikipedia, Google Search | 5(-) |
| UETQuintet(32) (Vietnam) | GPT-4o-mini | Adaptive routing: distill question → ensemble decides direct vs sequential. Direct: Google Search + Wikipedia TF-IDF lookup. Sequential: split into subquestions, retrieve/answer iteratively | Google Search, Wikipedia | 5(5) |
| PreceptorAI (33) Thailand | MedReason-8B, Gemini Pro 2.5 | Entity/phrase-driven retrieval: extract noun phrases → Wikipedia API summaries for phrase pairs → concatenate evidence → answer with MedReason-8B; fallback to Gemini if no evidence | Wikipedia | 4(4) |
| Fluxion (34) Saudi Arabia | Gemini 2.0/2.5 Flash | Prompting study: compare (a) zero-shot structured output prompt and (b) few-shot + CoT "expert" prompt | Wikipedia | 4(-) |
| CLaC (35) Canada | Qwen 2.5-Coder-32B Qwen 3-8B-AWQ | Two-system approach: (1) agentic decomposition → Wikipedia retrieval; (2) question splitting → retrieval from Wikipedia + PubMed → evidence assessment → answer distillation | Wikipedia, PubMed | 2(2) |
| Orekh. (36) Russia | Not specified in the text | Build custom health KB from Wikipedia → multi-hop decomposition → retrieve using BM25 + dense indexing → synthesize intermediate answers → final answer | Custom Wikipedia-derived KB | 3(-) |
| LasigeBioTM (37) Portugal | Gemma-3-1B-IT, MedGemma-27B, Mistral-7B-Instruct | Ontology + Wikipedia retrieval: extract entities (Mistral) → query MONDO + Wikipedia → concatenate evidence → final answer generation; plus baseline model benchmarking | Mondo Disease Ontology, Wikipedia | 4(2) |
| CaresAI (38) Australi | LLaMA 3 8B (fine-tuned + original) | Fine-tuning: train on dev set + supplemental QA → inference uses tuned model + a second prompt to extract concise answer; fallback to original model | Dev. set + suppl. QA pairs | 4(5) |
| BioHop (39) US | DeepSeek R1 | Plan-then-retrieve RAG: generate an "answer plan" → convert claims into subqueries → retrieve Wikipedia passages (Rag-Gym) + rerank → answer generation conditioned on plan + evidence | Wikipedia | 2(-) |

*Not all teams submitted additional runs.

Nine workshop papers were accepted to the MedHopQA track, as summarized in Table 2. All submitted systems were built around large language models (LLMs), and most employed a retrieval-augmented generation (RAG) architecture (15). Within this shared framework, system designs diverged primarily in their choice of model scale and openness, with some teams relying on proprietary frontier models and others adopting smaller open-source alternatives.

Two main optimization strategies emerged across the submissions: (i) retrieval- and agent-based pipelines that explicitly decompose and retrieve information before synthesis, and (ii) model-centric approaches that adapt the model itself through fine-tuning or data augmentation before generating concise answers.

Several teams—including DMIS Lab, UETQuintet, CLaC, Orekhovichi, and BioHop—treated the task primarily as a retrieval and composition problem. Their systems therefore implemented RAG-style pipelines or agent-based planning strategies with explicit query decomposition. These approaches typically generated sub-queries, performed iterative retrieval, and synthesized evidence across multiple hops. A lighter-weight RAG variant was used by the PreceptorAI and LasigeBioTM systems, which focused on extracting key entities from the question to retrieve targeted snippets from structured sources.

Nearly all teams used Wikipedia—the same source used to construct the dataset—as a primary knowledge base. Many systems implemented hybrid retrieval pipelines that combined lexical retrieval methods (e.g., BM25) with dense retrieval models to capture both keyword-based and semantic matches. Robustness was further improved through fallback strategies, such as routing queries from an open-source model to a larger proprietary model when the initial attempt failed.

In contrast, the CaresAI team pursued a model-centric approach, fine-tuning their model on the provided question–answer examples. In this system, prompting was used primarily for answer formatting and as a fallback mechanism rather than for driving the reasoning process itself.

**Selected Systems**

Below, we present six selected systems that illustrate the range of approaches used in the MedHopQA challenge.

**DMIS Lab team, Jongmyung Jung and Hyeongsoon Hwang, Korea University**

The DMIS Lab team (31) proposed a multi-step RAG framework with an explicit answer-selection stage. Their system combined three distinct retrieval strategies, followed by a decision-making module that selected the best final answer (40-42).

The first strategy, Query2Doc-based retrieval, used GPT-4o to generate a pseudo-document that expanded the original query. This expanded query was used to retrieve 1,000 documents from a Wikipedia corpus, which were then reranked with MedCPT to the top 200 before being passed back to GPT-4o. The second strategy used OpenAI o3 to generate a rationale-augmented query, followed by a similar retrieve-and-rerank pipeline implemented with ElasticSearch. The third strategy, web-search-augmented retrieval, used Gemini 2.5 Pro grounded with Google Search to capture information beyond the static corpus. Finally, a decision-maker model based on o3 evaluated the candidate answers produced by the three retrieval streams and selected the most accurate response, with a fallback mechanism that generated a new answer directly when none of the candidates was judged suitable.

**UETQuintet team, Quoc-An Nguyen, Thi-Minh-Thu Vu, Bich-Dat Nguyen, Dinh-Quang-Minh Tran, and Hoang-Quynh Le, VNU University of Engineering and Technology**

The UETQuintet team(32) proposed a RAG framework featuring selective decomposition, multi-source retrieval, and answer generation. Their system first simplified the input question using an LLM, then employed an ensemble of machine learning models to determine whether the question should be treated as direct or sequential. Questions classified as sequential were decomposed into sub-questions by GPT-4o-mini and linked in a multi-hop reasoning chain, whereas direct questions were handled as single-hop cases.

At each hop, the system retrieved relevant context from both the open web and Wikipedia using the current sub-question together with the answer from the previous hop. These retrieved snippets were then used for final answer generation by o3-mini. During the unofficial phase, the team introduced an additional refinement step in which contextual information was clustered to generate candidate answers, which were then presented to the LLM in multiple-choice format for final selection. The authors noted that the system showed strong potential even when built with relatively compact LLMs.

**CLaC team, Harikrishnan Gurushankar Saisudha, Ganesh Chandrasekar, University of Montreal**

The CLaC team(35) explored both agentic and non-agentic approaches to multi-hop biomedical question answering. All submitted systems used quantized Qwen-family LLMs together with RAG over Wikipedia and/or PubMed (43-45).

Their first system, Agentic-Qwen-Wikipedia, used a lightweight tool-calling agent from the smolagents library together with the Qwen2.5-Coder-32B-Instruct-GPTQ model for reasoning and planning. This system processed each question in a ReAct-style multi-step manner, iteratively generating sub-queries and retrieving information from Wikipedia until it determined that sufficient evidence had been collected.

To compare against this design, the team also developed LLM-Qwen-Wikipedia, a non-agentic system with an explicitly controlled iterative pipeline and memory module. In each iteration, the system generated a sub-question, extracted keyword-focused retrieval queries, retrieved supporting evidence, generated an intermediate answer, and then performed a termination check. The memory of previously answered sub-questions, together with the original question, was used to guide the next sub-question in the chain.

Their primary competition system, LLM-Qwen-Wikipedia-PubMed, extended this non-agentic pipeline by incorporating PubMed as an additional domain-specific retrieval source. The team reported that the non-agentic systems consistently outperformed the agentic variant, with the PubMed-integrated version achieving the strongest overall performance.

**OREKHOVICHI team, Rustam R. Taktashov, Nadezhda Yu. Bizyukova, Alexander V. Dmitriev, and Olga A. Tarasova, Institute of Biomedical Chemistry (IBMC)**

The OREKHOVICHI team (36) developed a fully local RAG-based system. They constructed a custom retrieval corpus by extracting more than 225,000 health-related articles from Wikipedia, using the PETScan Wikipedia Tool for category-based filtering. Their retrieval pipeline combined semantic vector search with keyword-based BM25S retrieval (46), followed by reranking using reciprocal rank fusion (47).

To address multi-hop questions, the system applied structured prompting to decompose each query into simpler sub-questions. These were processed independently by a local Llama3 med42 8B model (48), and the resulting intermediate answers were subsequently combined to produce a concise final response.

**lasigeBioTM team, Sofia I. R. Conceição and Paulo R. C. Lopes, LASIGE, Departamento de Informática, Faculdade de Ciências, Universidade de Lisboa**

The lasigeBioTM team(37) developed a RAG-based system using the Mistral-7B-Instruct-v0.3 model (49), augmented with external knowledge from Wikipedia and the Mondo ontology (50). They compared this configuration with the performance of MedGemma-27B (51).

Their experiments showed that although Mistral-7B-Instruct benefited from the addition of external knowledge, it did not match the baseline performance of MedGemma. In subsequent unofficial runs, the team found that pairing MedGemma with Wikipedia retrieval yielded only modest improvements, whereas combining MedGemma with structured information from the Mondo ontology produced their best results. These findings suggest that smaller models may struggle with complex domain-specific reasoning, while integrating domain-adapted models with structured biomedical knowledge, such as ontology data, can improve system accuracy.

**CaresAI team, Reem Abdel Salam, Mary Adewunmi, and Modinat Abayomi, Faculty of Engineering, Cairo University; Menzies School of Health Research, Charles Darwin University; Department of Biology, Boston College;**

The CaresAI team (38) adopted a supervised fine-tuning strategy based on LLaMA 3 8B, training the model on a curated biomedical question-answer dataset compiled from external sources including BioASQ, MedQuAD, and TREC. The team explored three experimental configurations: fine-tuning on combined short and long answers, fine-tuning on short answers only, and fine-tuning on long answers only.

To address challenges with overly verbose outputs, they also introduced a two-stage inference pipeline designed to produce concise short-answer responses. Although their models demonstrated some domain understanding and achieved stronger concept-level accuracy, their Lexical Match scores remained relatively low. The authors suggested that this discrepancy may reflect a gap between a model's semantic understanding and its ability to generate answers in the concise format required for automatic evaluation.

## Results and Discussion

*Table 3 Evaluation results ranked by the F1 score under the Lexical Match and MedCPT metrics. The best run is reported for each team. The table lists the median and mean values across all runs (in %).*

| Test Phase | | | |
|---|---|---|---|
| **Team Name** | **Best Run #** | **Lexical Match** | **MedCPT** |
| DMIS Lab | 2 | 87.3 | 89.3 |
| UETQuintet* | 5 | 83.8 | 86.0 |
| Insilicom | 3 | 80.0 | 83.1 |
| PreceptorAI* | 1 | 73.4 | 78.8 |
| Nckuiirlab | 3 | 66.9 | 76.7 |
| Fluxion | 1 | 68.1 | 72.0 |
| CLaC* | 2 | 67.6 | 71.8 |
| **Baseline** | **1** | **59.1** | **68.3** |
| Orekhovichi | 1 | 43.9 | 51.2 |
| Biojay | 1 | 45.8 | 48.8 |
| LasigeBioTM* | 4 | 28.3 | 34.2 |
| CaresAI* | 4 | 18.6 | 31.1 |
| LaosFun* | 3 | 26.1 | 29.4 |
| BioHop | 1 | 20.7 | 29.2 |
| **Median** | | **66.9** | **67.4** |

| Average | 54.65 | 58.2 |
| --- | --- | --- |

Table 2 summarizes the teams participating in the MedHopQA shared task, the systems they developed, and the number of submissions each team contributed. During the official test phase, each team was allowed up to five submissions. After reviewing the results and communicating with participants, we observed that some teams were constrained by resource limitations and could benefit from additional time to complete their experiments. To accommodate this, we opened an additional submission window, referred to as the "unofficial test phase," and invited all teams to submit additional runs. The subsequent submissions confirmed this expectation, as several teams improved both the number of questions answered and their overall scores.

Table 3 presents the evaluation results, including the best submission from each team, the baseline system, and the mean and median performance across all official runs. During the unofficial phase, the UETQuintet team improved their scores to 88.1 and 89.7 for lexical matching and concept-level evaluation, respectively. The lasigeBioTM team showed the most substantial improvement, reaching F1 scores of 56.2 and 61.4 for lexical and concept-level evaluation. Similarly, the CaresAI team demonstrated notable improvement, approximately doubling their initial scores.

Overall, the best-performing run achieved 89.3% accuracy, whereas the lowest-performing run achieved 0.3%, reflecting a submission that answered only a small fraction of questions. The baseline system achieved 68.3%, while the median performance across all runs was 67.4% and the mean was 58.2%. The difference between the mean and median indicates that several extremely low-performing runs substantially lowered the overall average. In total, 23 of 48 runs (48%) exceeded the baseline, suggesting that the baseline represented a competitive reference point rather than a trivial benchmark.

Performance stability varied markedly across teams. High-performing teams showed relatively tight clustering across their submitted runs (generally within a seven percentage-point range), whereas other teams exhibited much larger variation, with within-team spreads reaching approximately 30 percentage points. This variability suggests strong sensitivity to configuration choices such as prompting strategies, retrieval methods, normalization, and post-processing.

The results also suggest that model scale and retrieval-pipeline sophistication were important factors in system performance. The top-performing teams, DMIS Lab and UETQuintet, developed powerful but resource-intensive approaches using state-of-the-art proprietary models such as GPT-4o and Gemini as their primary reasoning engines. Their systems were further strengthened by multi-stage RAG pipelines that queried multiple external knowledge sources, including Wikipedia, and used advanced reranking algorithms to prioritize relevant evidence. This combination of a top-tier LLM and a robust multi-source retrieval proved effective for addressing the multi-hop reasoning required by the task.

In contrast, systems in the middle and lower tiers of the leaderboard explored a wider range of alternative strategies, often based on open-source models such as Mistral, LLaMA 3, and DeepSeek. These systems demonstrated methodological diversity, including fine-tuning on specialized biomedical datasets (CaresAI), incorporating structured ontology knowledge (lasigeBioTM), and applying reward-based training (BioHop). However, these approaches generally did not match the performance of larger proprietary models. This performance gap suggests that, for multi-hop biomedical reasoning tasks, the advanced reasoning capabilities and extensive knowledge encoded in large pre-trained models currently provide a substantial advantage over smaller models, even when the latter are augmented through fine-tuning or specialized knowledge sources.

### Performance by answer category

Table 4 analyzes performance by answer type and reveals substantial heterogeneity across categories. The baseline performed best on chemical (79.8%), species (81.8%), anatomical (76.2%), and yes/no (74.1%) questions, but was weaker on numerical (44.8%), gene/protein (55.5%), and chromosome (56.9%) questions. Across submitted runs, numerical questions were the main performance bottleneck: average accuracy was 22.5% (median 17.2%), and 12 of 48 runs scored 0% in this category, indicating frequent failures on tasks requiring exact numeric extraction or formatting. Chromosome-location questions were the second most difficult category, with an average accuracy of 47.2% and 5 of 48 runs scoring 0%. Yes/no questions showed unusually high variance, with a best-run score of 98.7%, compared with an average of 56.3% and a median of 74.4%, suggesting that calibration, negation handling, and evidence interpretation strongly influenced performance. After collapsing rare categories, the resulting “other” group (n=63) was comparatively easier for participants, with an average accuracy of 70.6%, and was the main category in which the greatest number of submitted systems outperformed the baseline (baseline 58.7%).

*Table 4 MedCPT evaluation results ranked by F1 score for each answer type category. For each team we selected their best run listed in Table 3. Highlighted values denote the best score in each category (in %).*

| **Team (Best)** | **Yes/ No** | **Dis** | **Gene / Prot** | **Chr** | **Che mDru g** | **Anat Part** | **Num** | **Sign/ Symp** | **Spec** | **Descr** | **Oth** |
|---|---|---|---|---|---|---|---|---|---|---|---|
| DMIS | 86.5 | **91.9** | **92.4** | **88.5** | **97.6** | **85.7** | **58.6** | 84.6 | **100** | **85.0** | **93.7** |
| UETQuin | **98.7** | 80.1 | 85.7 | 80.5 | 89.3 | 69.0 | 48.3 | 76.9 | 86.4 | 75.0 | 85.7 |
| Insilicom | 85.5 | 86.7 | 79.0 | 72.4 | 90.5 | 78.6 | 34.5 | **88.5** | **100** | 70.0 | **93.7** |
| Perc.AI | 82.8 | 81.0 | 75.6 | 72.4 | 89.3 | 78.6 | 31.0 | 65.4 | 77.3 | 75.0 | 82.5 |
| Nck.lab | 82.5 | 73.5 | 73.9 | 66.7 | 86.9 | 81.0 | 34.5 | 73.1 | 72.7 | 75.0 | 85.7 |
| Fluxion | 79.8 | 66.8 | 62.2 | 64.4 | 85.7 | 69.0 | 34.5 | 53.8 | 86.4 | 70.0 | 85.7 |
| ClaC | 82.5 | 68.7 | 65.5 | 65.5 | 81.0 | 54.8 | 37.9 | 53.8 | 81.8 | 75.0 | 69.8 |
| **Baseline** | **74.1** | **71.1** | **55.5** | **56.9** | **79.8** | **76.2** | **44.8** | **67.3** | **81.8** | **65.0** | **58.7** |
| Orekh. | 57.6 | 56.4 | 32.8 | 33.3 | 57.1 | 54.8 | 3.4 | 53.8 | 50.0 | 70.0 | 68.3 |
| Biojay | 1.0 | 73.0 | 65.5 | 54.0 | 85.7 | 83.3 | 13.8 | 61.5 | 72.7 | **85.0** | 73.0 |
| Lasige. | 15.8 | 45.0 | 26.9 | 25.3 | 58.3 | 47.6 | 10.3 | 34.6 | 54.5 | 45.0 | 69.8 |
| CaresAI | 57.2 | 20.9 | 24.4 | 0.0 | 31.0 | 9.5 | 0.0 | 23.1 | 31.8 | 25.0 | 31.7 |
| LaosFun | 33.3 | 23.7 | 29.4 | 29.9 | 25.0 | 23.8 | 13.8 | 26.9 | 31.8 | 35.0 | 44.4 |
| BioHop | 2.7 | 46.9 | 34.5 | 5.7 | 59.5 | 40.5 | 6.9 | 46.2 | 50.0 | 60.0 | 55.6 |
| Median (48 runs) | 74.4 | 69.9 | 65.1 | 57.5 | 82.1 | 69.0 | 17.2 | 61.5 | 77.3 | 75.0 | 81.0 |
| Average (48 runs) | 56.3 | 61.9 | 56.4 | 47.2 | 69.9 | 57.7 | 22.5 | 55.3 | 68.7 | 63.4 | 70.6 |

### Question-level consensus and notably difficult items

We quantified question difficulty by measuring the proportion of runs that answered each question correctly. Only 30 of 1,000 questions were solved by at least 90% of runs (≥44/48), and these were dominated by disease and gene/protein items, consistent with relatively unambiguous entity-lookup patterns. At the other extreme, 34 of 1,000 questions were solved by at most 10% of runs (≤4/48), with most concentrated in the numerical, gene/protein, and chromosome-location categories.

Notably, 16 questions were not solved by any submitted run (0/48), forming a persistent set of difficult items spanning mutation, gene/protein identification, precise chromosome-band notation, and some disease, anatomical, and sign/symptom questions. The baseline system, however, answered 4 of these 16 questions correctly, suggesting that at least some failures were due to normalization precision rather than a complete lack of relevant knowledge. For example, systems sometimes returned a gene instead of a specific mutation or produced slight deviations in chromosome-band notation. In contrast, yes/no questions did not appear in either extreme consensus group (≥90% or ≤10%), indicating that these items more often produced intermediate agreement across systems than uniform success or failure.

Overall, these results highlight (i) a substantial performance gap between the top and median systems, (ii) strong category-specific bottlenecks, particularly for numerical and chromosome-location questions, and (iii) a small but important set of persistently unsolved items that remain challenging for all submitted approaches.

## Conclusions

The MedHopQA shared task at BioCreative IX brings the following contributions: 1. A newly, manually created resource of multi-hop questions centered on diseases, chemicals, genes, and other aspects of medical information. 2. World-wide interest with the widest geographical spread of any of the previous BioCreative challenges.

Overall, these results highlight (i) a substantial performance gap between the top and median systems, (ii) strong category-specific bottlenecks, particularly for numerical and chromosome-location questions, and (iii) a small but important set of persistently unsolved items that remain challenging for all submitted approaches. The strongest-performing systems combined powerful proprietary models, including GPT-4o, o3-mini, and Gemini-2.5-pro, with carefully engineered retrieval and reasoning pipelines. Successful strategies included query reformulation, chain-of-thought prompting, multi-source retrieval, reranking, and controlled query expansion over domain-relevant corpora. High-performing teams also balanced effectiveness against practical cost constraints by avoiding unnecessarily large numbers of API calls. Successful runs involved semantic retrieval beyond NER, creative prompt engineering, model specialization and division of labor between models as well as controlled query expansion for domain corpora. More broadly, the results suggest that strong performance depends not only on the underlying model, but also on effective retrieval beyond simple entity matching, careful prompt design, specialization of different models for different subtasks, and disciplined orchestration of the overall pipeline.

The MedHopQA dataset remains available through Codabench and is intended to support continued development of systems capable of robust multi-hop reasoning in the biomedical domain.

## Acknowledgments

This research was supported by the Intramural Research Program of the National Institutes of Health (NIH). The contributions of the NIH authors are considered Works of the United

States Government. The findings and conclusions presented in this paper are those of the authors and do not necessarily reflect the views of the NIH or the U.S. Department of Health and Human Services.

The work of RRT and NYB is supported by the Program for Basic Research in the Russian Federation for a long-term period (2021-2030) (№ 124050800018-9).

The work of JJ and HH is supported by the National Research Foundation of Korea (NRF-2023R1A2C3004176), and the Korea Health Technology R&D Project through the Korea Health Industry Development Institute (KHIDI), funded by the Ministry of Health & Welfare, Republic of Korea (RS-2022-KH129295).